\pdfoutput=1

\documentclass[11pt]{article}

\usepackage[]{acl2023}

\usepackage{times}
\usepackage{latexsym}

\usepackage{graphicx}
\usepackage[nolist]{acronym}
\usepackage{algorithm}
\usepackage{algorithmic}

\begin{acronym}
    \acro{rq}[RQ]{research question}
    \acroplural{rq}[RQs]{research questions}
    \acro{nlp}[NLP]{natural language processing}
    \acro{plm}[PLM]{pretrained language model}
    \acroplural{plm}[PLMs]{pretrained language models}
    \acro{ai}[AI]{artificial intelligence}
    \acro{qa}[QA]{Question answering}
    \acro{kbqa}[KBQA]{question answering over knowledge bases}
    \acro{nlg}[NLG]{Natural language generation}
    \acro{kg}[KG]{knowledge graph}
    \acroplural{kg}[KGs]{knowledge graphs}
\end{acronym}

\usepackage[T1]{fontenc}

\usepackage[utf8]{inputenc}

\usepackage{microtype}

\usepackage{inconsolata}

\title{From Data to Dialogue: Leveraging the Structure of Knowledge Graphs for Conversational Exploratory Search}

\author{Phillip Schneider$^1$, Nils Rehtanz$^1$, Kristiina Jokinen$^2$ and Florian Matthes$^1$ \\
     $^1$Technical University of Munich, Department of Computer Science, Germany \\
     $^2$AI Research Center, National Institute of Advanced Industrial Science and Technology, Japan \\ 
     \texttt{\{phillip.schneider, nils.rehtanz, matthes\}@tum.de} \\
     \texttt{kristiina.jokinen@aist.go.jp}}

\begin{document}
\maketitle
\begin{abstract}
Exploratory search is an open-ended information retrieval process that aims at discovering knowledge about a topic or domain rather than searching for a specific answer or piece of information. Conversational interfaces are particularly suitable for supporting exploratory search, allowing users to refine queries and examine search results through interactive dialogues. In addition to conversational search interfaces, knowledge graphs are also useful in supporting information exploration due to their rich semantic representation of data items. In this study, we demonstrate the synergistic effects of combining knowledge graphs and conversational interfaces for exploratory search, bridging the gap between structured and unstructured information retrieval. To this end, we propose a knowledge-driven dialogue system for exploring news articles by asking natural language questions and using the graph structure to navigate between related topics. Based on a user study with 54 participants, we empirically evaluate the effectiveness of the graph-based exploratory search and discuss design implications for developing such systems.
\end{abstract}

\section{Introduction}
Dialogue systems, or conversational interfaces, have been a long-established research area in \ac{nlp}. Over several decades, their capabilities have continuously improved, and due to recent advances in neural network-based methods, they have progressed even further. Today, we see widespread adoption of dialogue systems across a plethora of domains, including customer service, education, healthcare, home automation, and finance, to name a few \cite{maedche2019ai}. There is currently significant public interest in auto-regressive large language models, such as PaLM or GPT-3 and its successors ChatGPT and GPT-4 \cite{brown2020language,chowdhery2022palm,openai2022chat}. These generative models are fine-tuned with instructions and can produce human-like responses for a variety of prompts, but they are susceptible to generating text that lacks factual grounding in credible information sources. Nevertheless, this trend is another driving force behind the paradigm shift in moving human-machine interactions toward the medium of language. 

\begin{figure}[t]
  \centering
  \includegraphics[width=0.99\linewidth]{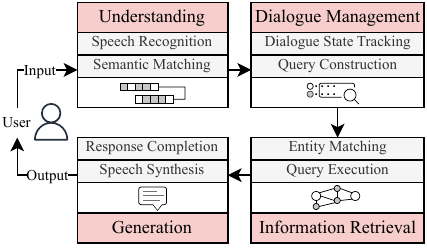}
  \caption{Conceptual architecture of the exploratory search system.}
\label{fig:architecture}
\end{figure}

Although most dialogue systems can be categorized as either task-based, performing specific functions like controlling smart devices, or non-task-based, engaging in social chit-chat, another use case that is gaining research attention is the application of conversational interfaces for information retrieval. While traditional search systems are limited to single-turn keyword query-response or question-answer interactions, conversational search systems support multi-turn information-seeking dialogues \cite{radlinski2017theoretical}. Hence, conversational interfaces are particularly suitable for exploratory search settings, in which information seekers have a vague intention of what they are looking for because they are unfamiliar with a complex and multifaceted topic \cite{white2009exploratory}. Conversational search systems enable users to refine their queries iteratively, evaluate search results, give feedback, and narrow down the information space. This intuitive, dialogue-based interaction theoretically promises to overcome long-known information retrieval issues like vocabulary mismatch or the anomalous state of knowledge, which restricts users from forming meaningful queries without prerequisite knowledge in an unfamiliar domain \cite{belkin1980anomalous,furnas1987vocabulary}.

Knowledge graphs can capture the interconnected nature of domain-specific knowledge and have proven to be a powerful data structure for dialogue systems and numerous \ac{nlp} tasks \cite{shaoxiong2022survey,schneider-etal-2022-decade}. As a rich representation of entities and semantic relationships, scholars have studied them to improve dialogue systems by enabling contextual utterance understanding, facilitating decision-making, or enhancing retrieval-based response generation. There are several reasons why knowledge graphs are suitable for exploratory search. For example, knowledge graphs can help with the disambiguation of complex search queries, which is helpful for topics where the meaning of terms may vary depending on context. In addition, the graph structure allows users to navigate through related concepts and explore different facets of a topic. Because these connections approximately reflect associative thinking patterns of human cognition, they can aid serendipitous discovery and novel insights into topics \cite{bahareh2014exploring}. 

Despite the growing body of research on combining conversational interfaces and knowledge graphs, there is an apparent lack of studies focusing on exploratory search. While a few studies mention the theoretical potential of these technologies for browsing and discovery, the effectiveness of implemented dialogue systems in real-world settings has yet to be studied thoroughly. To address this gap, we developed a knowledge-driven conversational system as a baseline and evaluated it in-depth with a sample of 54 users. We opted for the application domain of news search because it involves browsing numerous topics that change daily, as elaborated upon in \citet{schneider2023voice}. At the time of this study, there are no existing commercial or scientific solutions available that support conversational news exploration. Our main contributions are three-fold:
\begin{enumerate}
    \item We propose exploratory search mechanisms based on the structure of knowledge graphs.
    \item We developed a conversational search system for exploring a graph with news articles.
    \item We empirically evaluated the developed prototype and publish our source code and collected data in a GitHub repository.\footnote{\href{https://github.com/sebischair/kg-conv-exploratory-search}{github.com/sebischair/kg-conv-exploratory-search}}
\end{enumerate}


\section{Related Work}
\label{sec:related-work}
Integrating knowledge graphs into dialogue systems has been a subject of interest within the research community for many years. Knowledge graphs have the potential to enhance utterance understanding, response generation, and dialogue management. Focusing on the first area, \citet{zhou2020improving} employ word- and concept-oriented graphs to better understand short user utterances through contextual enrichment. The authors demonstrate with experiments that their model outperforms existing approaches on recommendation tasks. Similarly, \citet{wang2020improving} propose TransDG, a model transferring abilities of question representation and knowledge matching from the task of knowledge-based question answering to improve utterance understanding and response generation. Numerous other studies have investigated using structured information in graphs to ground responses generated by language models \cite{chen2019kbrd,zhang2020grounded,chaudhuri2021grounding}. 

While understanding and generating dialogues play crucial roles, knowledge graph-based dialogue management is especially relevant to conversational search because it is responsible for handling information processing tasks. In a study from \citet{sarkar-etal-2020-suggest}, a movie recommendation system is proposed that generates suggestions from constructed subgraphs by considering dialogue states and user profiles. Another closely related work is WikiTalk from \citet{wilcock-2012-wikitalk} who introduces a dialogue-based knowledge access system for Wikipedia articles. Although the system does not rely on a knowledge graph with semantically related entities, it uses hyperlinks extracted from articles to transition between topics.

Diverging from previous studies, our research explicitly focuses on the phenomenon of exploratory search, where users are confronted with unfamiliar territories of information \cite{schneider2023investigating}. To the best of our knowledge, we are the first to conduct a practical evaluation of how knowledge graphs can be exploited to address the unique challenges posed by exploratory search, providing a first baseline system. Our proposed voice-based conversational agent leverages the semantic graph structure to compress the search space through entity-centric filtering within a targeted scope and expand the search space by suggesting related entities, familiarizing users with the knowledge domain, and offering opportunities for further exploration beyond their initial query.

\section{Knowledge Graph and Conversational Search System Architecture}
\label{sec:system-architecture}

\subsection{Knowledge Graph Construction and Graph-Based Search}
\begin{figure}[h]
  \centering
  \includegraphics[width=0.7\linewidth,trim=0.1mm 5mm 5mm 1mm,clip]{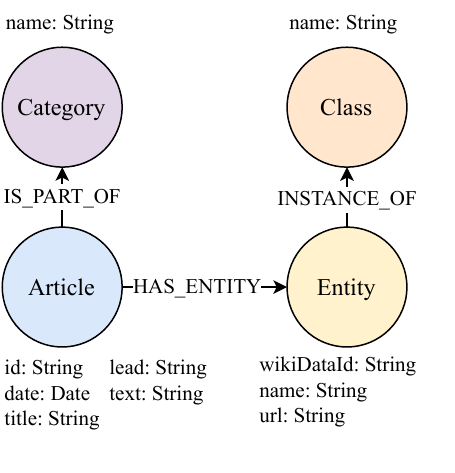}
  \caption{Graph structure of semantic data model.}
  \label{fig:data-model}
\end{figure}

To create the graph-based exploratory search system, we gathered a news article corpus from the Tagesschau news program.\footnote{\href{https://tagesschau.api.bund.dev}{Tagesschau API: https://tagesschau.api.bund.dev}} We automatically scraped and organized all news articles into a knowledge graph structure. Concerning the data model displayed in Figure~\ref{fig:data-model}, there are distinct node types for articles, news categories, and entities (e.g., Barack Obama), the latter being instances of entity classes (e.g., human). Each article is part of one news category, such as politics or sports, but can be linked to multiple entity nodes. To derive entities mentioned in a given article text, we performed named entity recognition and linking to the target knowledge base Wikidata.\footnote{\href{https://www.wikidata.org}{Wikidata knowledge base: https://www.wikidata.org}} We make use of a wikification model from \citet{brank2017annotating}, which supports multiple languages and provides language-agnostic Wikidata identifiers for the graph, making it feasible to adapt the system to other languages, such as Chinese, Spanish, or English. 

Over a period of nearly two months, 2118 German news articles were continuously added to the knowledge graph, which comprises a total of 6262 nodes and 19125 relationships. Additional information about the number of nodes and relationships can be found in Figure~\ref{fig:news-kg} in Appendix~\ref{sec:appendix}.

\begin{figure*}[!h]
  \centering
  \includegraphics[width=0.95\linewidth]{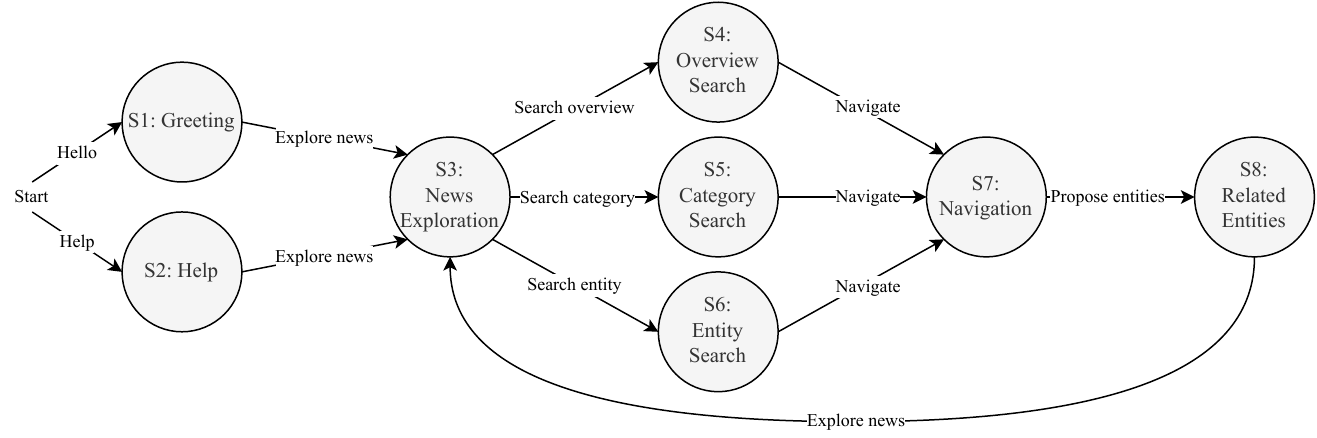}
  \caption{Finite-state model of dialogue states for conversational exploratory search system.}
\label{fig:finite-state-machine}
\end{figure*}

Building upon the constructed news knowledge graph, we can leverage graph-based search mechanisms for the dialogue system. Figure~\ref{fig:finite-state-machine} illustrates the possible information-seeking paths as a finite-state machine with eight states (S1-S8). First, a user can initiate the conversation by greeting the agent (S1) or asking for help (S2). Thereafter, the agent introduces three search mechanisms (S3), namely, getting an overview of recent popular articles (S4), searching articles by news category (S5), and entity-based search (S6). We opted to incorporate these three levels of abstraction in order to cover varying degrees of specificity. The agent queries the graph to retrieve articles relevant to the user's search intent and suggests three articles by listing their titles. Next, the user can navigate (S7) by either asking for additional articles or selecting a specific one. The navigation options also include the ability to skip, stop, or repeat articles being read at the moment. Finally, when the agent has finished reading a selected article, it suggests related entities (S8) as a possibility to explore more themes. This enables users to acquire knowledge about existing entities, discover related content, and deepen their understanding of a specific topic.

For each of the eight dialogue states, we defined intents and related entities along with training phrases that were used to fine-tune pre-trained intent detection and entity extraction models from Google. To suggest related entities for a given news article, a count-based suggestion algorithm was used, which is described below in pseudo-code.

\begin{algorithm}
\caption{Entity Suggestion Algorithm}
\small
\begin{algorithmic}[1]
  \REQUIRE news article $A$, knowledge graph $G$
  \STATE $linked\_entities \gets$ get\_article\_entities($A$, $G$)
  \STATE $entity\_freq \gets$ init\_empty\_dict()
  \FOR{$e$ in $linked\_entities$}
  \STATE $entity\_freq[e] \gets$ get\_article\_count($e$, $G$)
  \ENDFOR
  \STATE $sorted\_entities \gets$ sort\_by\_freq($entity\_freq$)
  \STATE $result\_entities \gets$ get\_top\_n($sorted\_entities$, $n$)
  \RETURN list of $n$ suggested entities $result\_entities$
\end{algorithmic}
\end{algorithm}

\subsection{Architecture of Conversational Search System}
The system architecture, depicted in Figure~\ref{fig:architecture}, consists of four core components for natural language understanding, dialogue management, information retrieval, and natural language generation. It is implemented as a web application using open-source Python libraries and Google Cloud models. Since the interaction happens through a voice interface, we use pre-trained speech-to-text and text-to-speech models from Google. The front-end is a web application with a single button for activating the microphone. When a user initiates the conversation by clicking the button, the recorded audio is transcribed with the speech-to-text model and sent to the natural language understanding component, which is built with the conversational agent framework Google Dialogflow.\footnote{\href{https://cloud.google.com/dialogflow}{Google Dialogflow: https://cloud.google.com/dialogflow}} This framework is responsible for producing a semantic representation of the user input by classifying intents, recognizing entities, and keeping track of the dialogue state. 

The dialogue management, which uses the dialogue state and context to determine the system's actions, is implemented as a Python-based conversation fulfillment service with Flask. After matching the semantic representation of the user input to query templates, the dialogue manager uses the extracted entities and links them to entities present in the knowledge graph to construct graph queries. The knowledge graph itself is stored as a property graph in a Neo4j database.\footnote{\href{https://neo4j.com/}{Neo4j graph database: https://neo4j.com/}} The queries are executed, and the results are combined with a text template in a probabilistic manner to complete a response. Lastly, the text-to-speech model generates an audio file from the response text that is automatically played back to the user. Despite the system's various technical components depicted in Figure~\ref{fig:architecture-big}, the voice interface hides the complexity of the underlying graph-based information retrieval, providing a highly accessible search experience. 

\section{System Evaluation Results}
\label{sec:results}
The goal of the evaluation was to determine the effectiveness of the proposed graph-driven dialogue system for exploratory search. To achieve this, we conducted an extensive human evaluation focused on the example scenario of a news search. Collecting empirical observations helped to reveal insights into technical limitations and how the conversational search system was utilized by the study participants in a practical setting.

\subsection{Study Sample}
A diverse sample of participants was acquired by recruiting test users through university courses, social media platforms, and snowball sampling to expand the pool of test users further. The voice-based dialogue system was made available through a minimalist web interface with one button to activate the microphone. Participants received no concrete instructions on how to use the agent or the existing news articles since we wanted to assess the conversational agent's self-explanatory capabilities. They were only told to greet the agent, seek information about news, and afterward report their experience in an online questionnaire. Additionally, we logged all dialogue interactions between the conversational agent and the participants.

The gender distribution among the study's participants was fairly balanced, with 43\% being female and 57\% male. The participants' ages ranged from 14 to 86, encompassing different age groups. The mean age was 36, and young adults between 20 and 30 accounted for the biggest group. We also inquired about the participants' familiarity with virtual assistants. Of those surveyed, 76\% of respondents had previous experience, but those above 50 had considerably less experience. Even though almost all participants consumed daily news, and many were familiar with virtual assistants, only 2 out of 54 had used virtual assistants to ask about the news. Instead, most respondents preferred other channels, such as the Internet, radio, television, or physical newspapers.

Consequently, the evaluation study included a diverse sample of participants across different age groups and genders, where most never had used conversational agents for news search. This renders our participant pool an ideal foundation for evaluating how accessible and effective the proposed conversational exploratory search system is.

\subsection{Conversation Log Analysis} 
An important part of our empirical evaluation involved logging and analyzing the dialogues between the conversational agent and test users. Over a span of two weeks, 54 participants acted as test users of the exploratory news search system. Because a few users started more than one conversation, we collected 65 dialogues with a total sum of 2172 dialogue turns, where one turn refers to a single utterance by either the agent or user. On average, each dialogue session comprised 33.4 turns. The mean duration of a dialogue session was around 9 minutes and 11 seconds, with a standard deviation of 7.8 minutes. The longest recorded conversation lasted for roughly 41 minutes, and the shortest for 22 seconds.




Concerning the three different search intents, entity-based search was by far the most commonly detected intent, with a share of 10.2\% of all dialogue turns, followed by searching for a general overview and searching for news categories with 2.2\% and 0.9\%, respectively. Navigating intents, such as selecting or skipping articles, were also frequently used, making up 15.8\% of the turns. Whenever the agent could not determine the intent of a user's utterance, it replied with a fallback clarification response. To give an example, some users asked for information about the weather or about what a specific person did on a particular day, although we did not include intents for these types of requests. In other cases, issues arose due to nonsensical audio transcripts generated by the speech-to-text model. Overall, the fallback response appeared 192 times in the logs, accounting for a share of 8.8\%. 

Since entity-based search proved to be the most popular search mechanism in the conversation logs, we examined which specific entities users were looking for. Table~\ref{tab:entity-matching-performance} lists the number of the top three recognized and unrecognized entity classes, as well as their accuracy scores, indicating how often the conversational agent could match them correctly to the graph entities. Of the 222 entity-related search requests in the logs, the conversational agent successfully recognized 130 entities (59\%) and linked them to entities in the news knowledge graph. Entities of the class country appeared most often and were recognized with a very high accuracy of 94\%. Similarly, classes like person or city also had high accuracy scores of at least 70\%. Considering the unrecognized entities, we see that entities of the classes sports, person, and nutrition were among the most common, although they constituted a minority compared to the long tail of other classes. There were numerous topics that users searched for that were not covered by the Tagesschau news portfolio. These included news about celebrities, computer games, local events, and weather reports. A detailed discussion of reasons regarding issues with unrecognized entities is part of Section~\ref{sec:discussion}. 

\begin{figure*}[h]
  \centering
  \includegraphics[width=0.96\linewidth]{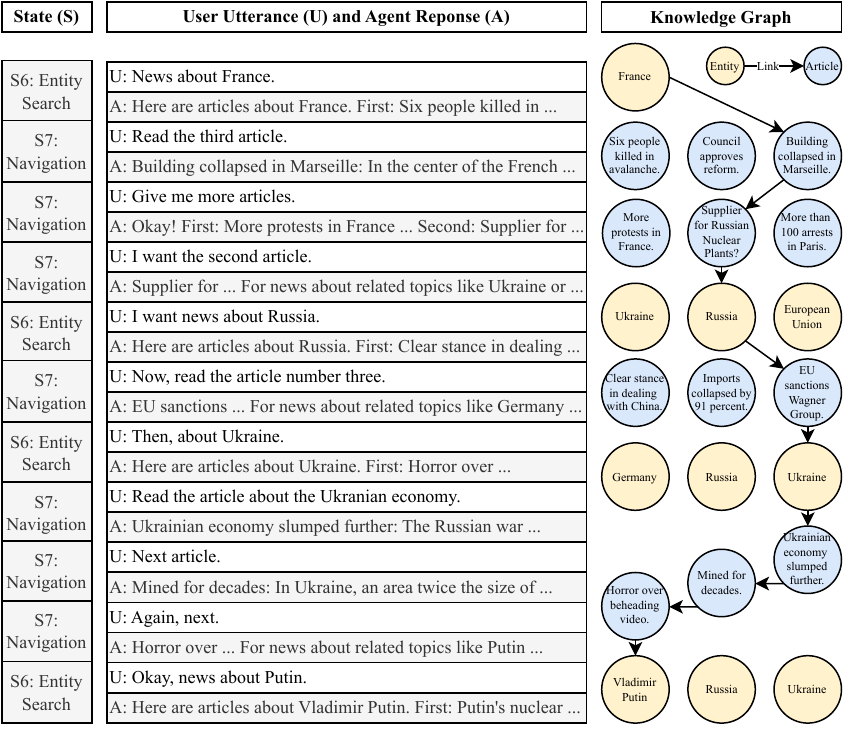}
  \caption{Visualization of entity-centric search history in constructed news knowledge graph.}
\label{fig:search-path}
\end{figure*}

\begin{table}[h]
\small
\centering
\begin{tabular}{lcc}
\hline
\textbf{Entity Types} & \textbf{Count} & \textbf{Accuracy}\\
\hline
Recognized Entities & 130 & 0.59 \\
\hspace{1em} Country & 53 & 0.94 \\
\hspace{1em} Person & 19 & 0.70 \\
\hspace{1em} City & 14 & 0.78 \\
\hspace{1em} Other classes & 44 & - \\
\hline
Unrecognized entities & 92 & 0.41 \\
\hspace{1em} Sports & 10 & 0.17 \\
\hspace{1em} Person & 8 & 0.70\\
\hspace{1em} Nutrition & 7 & 0.29\\
\hspace{1em} Other classes & 67 & - \\
\hline
\end{tabular}
\caption{Count and accuracy of recognized and unrecognized entity classes.}
\label{tab:entity-matching-performance}
\end{table}

Our underlying hypothesis regarding the combination of knowledge graphs and conversational agents is that the connections within the graph structure can be leveraged for exploration purposes. In theory, human thinking patterns revolve around mental associations between concepts \cite{morewedge2010associative}, which knowledge graphs can model as directed edges between two entities. Through our comprehensive system evaluation, we gathered numerous dialogues that serve as evidence of participants intuitively using the suggested linked entities to explore related articles on a given subject. To illustrate this point, Figure~\ref{fig:search-path} shows one exemplary conversation transcript translated from German to English. The corresponding knowledge graph entities are shown next to the conversation logs.

The dialogue begins with the user's news request about France. In response, the system uses entity-based search to retrieve three articles linked to the entity France. The user chooses the third headline, and the agent reads the article on a collapsed building in Marseille. Then, the user expresses the desire to get more articles, leading the system to provide three additional articles related to France. Among these, the user selects the second article focusing on a French company's plans to supply nuclear fuel rods for Russian power plants. Subsequently, the agent's algorithm suggests the three most relevant entities mentioned in the article: Ukraine, Russia, and European Union. At this point, the user wants news about Russia and opts for an article concerning EU sanctions against the Russian Wagner Group. Following this, the agent once again suggests three entities and the user requests news related to the country of Ukraine. After skipping two articles, the user listens to the third one about a beheading video of a Ukrainian soldier and receives three related entities again. Continuing the exploratory news search, the user requests news about Russian President Vladimir Putin.


\subsection{Human Evaluation}
After the participants tested the conversational news exploration, they were asked to complete an evaluation questionnaire. The goal was to document their usage experiences and perceived satisfaction with the system. The questionnaire comprised both open-ended and closed-ended questions. For example, it included free-form text fields to share experienced issues and suggestions for improvement. In addition, questions about overall satisfaction, quality of recommendations, the relevance of information content, and other aspects were asked. These were rated on a Likert scale from 1 to 5, with 1 representing ''very poor'' and 5 ''very good''. Furthermore, participants were asked questions from the System Usability Scale (SUS), an established method developed to determine the usability of an application \cite{brooke1996sus}. In this, ten questions are asked on a 5-point Likert scale, ranging from 1 (''strongly disagree'') to 5 (''strongly agree''). The SUS score is calculated using a mathematical formula, and its resulting value ranges from 0 to 100, where a value of 68 is considered average. It is important to note that the SUS scores of two given applications are not directly comparable.

Table~\ref{tab:evaluation-questionnaire} lists the mean and standard deviation for the evaluation criteria of the questionnaire. Respondents rated their overall satisfaction score at 3.7, indicating a generally positive perception of the conversational search agent. The relevancy of the provided information received a mean value of 3.8 and strongly correlated with the satisfaction level. Ratings for the control over the dialogue and the agent's suggestions were also favorable, with scores of 3.7 and 3.8, respectively. While respondents rated the agent's capability to correctly understand requests a bit lower at 3.5, they appreciated the agent's comprehensible responses, resulting in the highest average rating of 4.4. In contrast, the perceived human-likeness of the agent got a mean value of 2.8, which is the lowest rating. Regarding the conversational search system's usability, we measured a SUS score of 79.4, surpassing the considered average of 68. It is notable that the age of participants had a significant influence on perceived usability, with those below the sample mean age of 36 giving an average SUS score of 86, compared to 68.3 for participants above 36. This aligns with the observation that older participants reported more difficulties using the system, which was reflected in lower general satisfaction scores. Besides the mentioned correlation between information relevance and overall satisfaction, the latter was also positively correlated with both the agent's suggestion and the control over the dialogue. In our analysis, the questionnaire data showed neither implausible correlations nor unexpected outliers, indicating that the users had no problems understanding and answering the questions.   

\begin{table}[h]
\small
\centering
\begin{tabular}{lcc}
\hline
\textbf{Evaluation Criteria} & \textbf{Mean} & \textbf{SD}\\
\hline
Overall satisfaction & 3.7 & 1.2 \\
Information relevancy & 3.8 & 1.1 \\
Control over dialogue & 3.7 & 1.2 \\
Understanding of requests & 3.5 & 1.3 \\
Comprehensible responses & 4.4 & 0.9 \\
Human-likeness of agent & 2.8 & 1.1 \\
Suggestions of agent & 3.8 & 1.0 \\
\hline
System Usability Scale (SUS) & 79.4 & 18.5 \\
SUS (age < 36) & 86.0 & 13.3 \\
SUS (age > 36) & 68.3 & 21.0 \\
\hline
\end{tabular}
\caption{Overview of evaluation criteria from the questionnaire with mean and standard deviation (SD).}
\label{tab:evaluation-questionnaire}
\end{table}


\begin{table*}[h]
    \small
\centering
  \begin{tabular}{lp{6cm}p{6cm}}
    \hline
    \textbf{Issue Group} & \textbf{Description} & \textbf{Example Dialogue Excerpt} \\
    \hline
    \small Acronyms & \small Entities were misrecognized as acronyms, but recognized correctly when spelled out in full. & \small User: News about ''BVB'' (football club). \newline Agent: Sorry, I cannot find news for this entity.\\
    \hline
    \small Disambiguation & \small While entities were correctly identified, they were sometimes recognized within the wrong context in cases of word ambiguity. & \small User: Give me news about the ''Wolf'' (animal). \newline Agent: Here is news about ''Wolf'' (article of person with surname Wolf). \\
    \hline
    \small Foreign language & \small Entities in a foreign language were often incorrectly transcribed by the speech-to-text model. & \small User: I want news about ''chaditip'' (ChatGPT). \newline Agent: Here is news about ''Charité'' (hospital).\\
    \hline
    \small Fuzzy matching & \small Sometimes, using fuzzy matching resulted in linking to wrong entities in the knowledge graph. & \small User: News about ''Justin Bieber'' (person). \newline Agent: Here is news about ''Biber'' (animal). \\
    \hline
    \small Multiple entities & \small If two or more entities were mentioned simultaneously, only one entity was recognized. & \small User: News about ''China'' and ''Taiwan'' (city). \newline Agent: Here is news about ''China'' (city). \\
    \hline
\end{tabular}
\caption{Identified issues regarding entity recognition and linking in exploratory search dialogues.}
  \label{tab:entity-recognition-issues}
\end{table*}

\section{Discussion}
\label{sec:discussion}
Our proposed conversational search system was built upon the hypothesis that combining conversational interfaces and knowledge graphs can support exploratory information-seeking. In order to examine the synergistic effects between these two technologies, we developed a baseline system for the application domain of exploring news articles. We conducted an extensive user study with a diverse pool of participants to evaluate the system, gathering empirical observations of how the test users interacted with the conversational agent to explore the underlying news knowledge graph. 

One of the key insights from the questionnaire results was that participants generally enjoyed using the system. They found it easy to understand the system's functionalities and expressed a strong liking for its intuitive control through voice commands. This was reflected in a high SUS score of 79.4, indicating a favorable user experience. In the questionnaire's free-text responses, participants emphasized the accessibility of the voice interface, noting its benefits for disabled individuals and situations where typing is impractical, such as when driving a car. Besides, the feedback revealed other aspects that contributed to the overall positive reception of the system. In particular, users appreciated the freedom to search among a plethora of topics and select from related suggestions relevant to their interests. These findings support our hypothesis that the underlying knowledge graph structure can facilitate the selection of relevant information and enhance the information-seeking experience.

However, we also identified certain issues encountered by some users regarding the search-centered dialogue. A subset of participants reported facing technical difficulties with their microphones, which coincided with problems with the agent's understanding of requests and inaccurate retrieval of articles. Interestingly, these issues were more commonly mentioned by older users above the age of 50, who tended to give worse usability scores, too. We assume that these users might have had lower technical literacy and were less familiar with using voice assistants. This observation aligns with related studies involving elderly individuals \cite{pradhan2020use,kim2021exploring}. Additionally, respondents expressed a desire to have more news sources than solely the Tagesschau because they were interested in topics about fashion, music, and local events. Furthermore, participants suggested improvements in the speech generation of the agent. For instance, they proposed incorporating more diverse responses and enhancing the pronunciation of the German text-to-speech models, especially for English terms. 

Another important finding pertains to the effectiveness of entity-based search for discovering news. Despite participants initially having no knowledge about available articles, they intuitively requested news for specific entities of interest. Then, the agent's suggestions of related entities played a crucial role in reinforcing the exploration of related topics. We have logged multiple dialogues that demonstrate this navigation between entities and linked news articles. One concrete example is shown in Figure~\ref{fig:search-path}. The dialogue transcript provides tangible evidence of how conversational agents can leverage the semantic graph structure of interlinked entities and a finite set of dialogue states for open-ended exploratory search, helping users overcome the hurdle of not knowing precisely what to ask by offering topics for further exploration.  

On one side, our analysis revealed a good performance on correctly matching entities from certain classes, such as cities, countries, and persons, which were also the most frequently requested (see Table~\ref{tab:entity-matching-performance}). On the other side, searching for entities was inherently prone to issues due to the virtually infinite number of possible entity-specific queries, whereas the knowledge graph contained only 2874 entities. To gain a better overview of the problems regarding entity matching in exploratory search dialogues, we identified five issue groups as presented in Table~\ref{tab:entity-recognition-issues}. Firstly, when entities were mentioned as acronyms, they were usually misrecognized until they were spelled out in full. Secondly, despite entities being correctly identified, there were a couple of cases where they matched within the wrong context, particularly in situations with word ambiguity. Thirdly, entities expressed in foreign languages often suffer from transcription errors by the speech-to-text model, resulting in recognition inaccuracies. Fourthly, using fuzzy matching occasionally led to the incorrect matching of entities from the article to the knowledge graph. Lastly, when multiple entities were mentioned simultaneously, only one entity was recognized, potentially omitting relevant information. These findings shed light on the conversational exploratory search system's current limitations and point out areas for improvement in entity recognition and linking.

\section{Conclusion}
\label{sec:conclusion}
In this study, we propose a conversational exploratory search system with a news knowledge graph. It leverages the graph structure to retrieve relevant news articles and suggest related ones. We demonstrate the effectiveness of this baseline system through an extensive human evaluation, providing evidence that participants used the underlying semantic relations to explore articles across various domains. We discuss not only identified challenges with entity recognition and linking but also technical issues regarding speech recognition and generation. These limitations serve as design implications for building more advanced systems and enable meaningful comparisons with our baseline. Our future work will concentrate on techniques to address challenges associated with accurately matching entities from user requests to the knowledge graph in multiple languages. Moreover, we plan to integrate large language models for enhanced response generation while using the graph structure as explicit knowledge grounding. By doing so, it is our aim to generate more natural, contextually appropriate responses and augment the search experience with personalized news summaries.

\section*{Acknowledgements}
This work has been supported by the German Federal Ministry of Education and Research (BMBF) Software Campus grant 01IS17049. Kristiina Jokinen acknowledges the support of Project JPNP20006 commissioned by the New Energy and Industrial Technology Development Organization (NEDO), Japan.

\bibliography{anthology,custom}
\bibliographystyle{acl_natbib}

\newpage
\onecolumn
\appendix
\section{Appendix}
\label{sec:appendix}
\begin{figure*}[h]
  \centering
  \includegraphics[width=0.75\paperwidth]{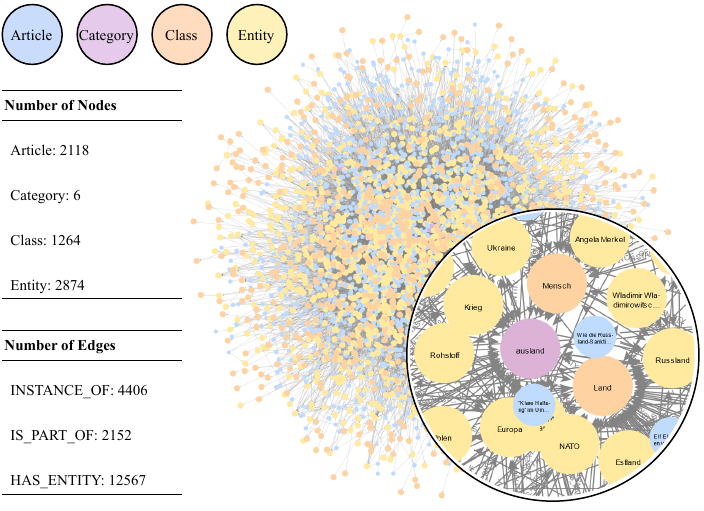}
  \caption{Overview of the different node and edge types in the automatically constructed news knowledge graph.}
\label{fig:news-kg}
\end{figure*}

\begin{figure*}[h]
  \centering
  \includegraphics[width=0.75\paperwidth]{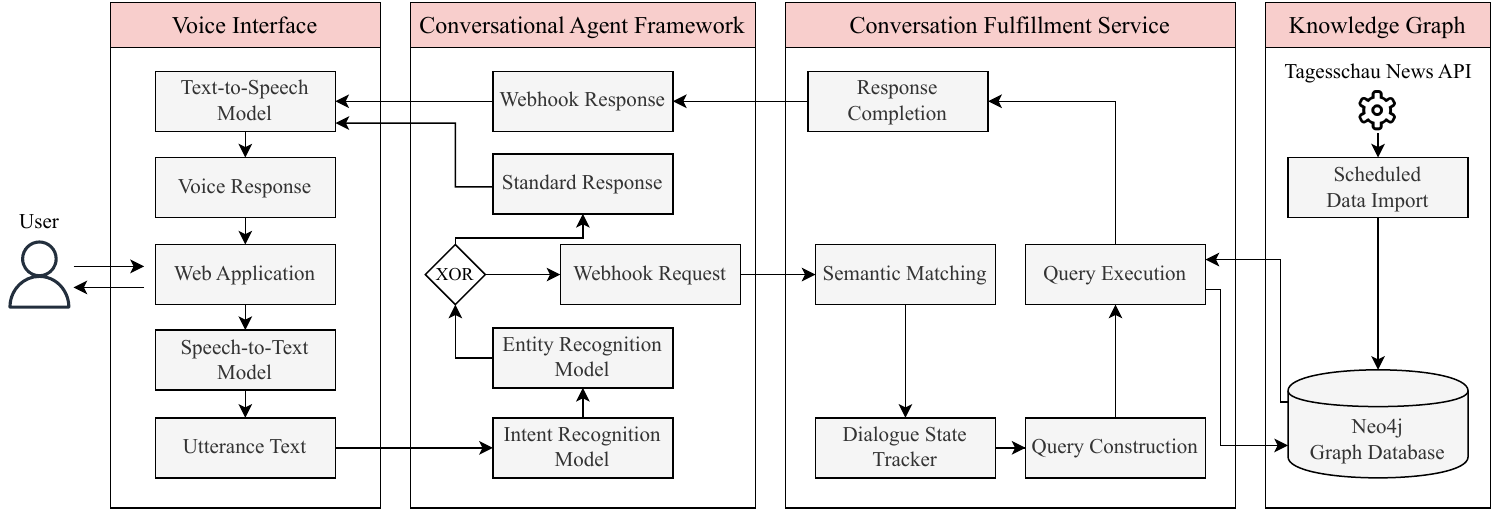}
  \caption{Architectural components of the proposed conversational exploratory search system.}
\label{fig:architecture-big}
\end{figure*}

The appendix provides further insights into the results of our research, including an overview of the news knowledge graph and the architectural components of the conversational exploratory search system, as depicted in Figures~\ref{fig:news-kg} and \ref{fig:architecture-big}, respectively.

\end{document}